\documentclass{article}
\usepackage{ijcai25}
\usepackage[T1]{fontenc}    % use 8-bit T1 fonts

\usepackage{times}
\usepackage{soul}
\usepackage{url}
\usepackage[hidelinks]{hyperref}
\usepackage[utf8]{inputenc}
\usepackage[small]{caption}
\usepackage{graphicx}
\usepackage{amsmath}
\usepackage{amsthm}
\usepackage{booktabs}
\usepackage{algorithm}
\usepackage{algorithmic}
\usepackage[switch]{lineno}
\usepackage{amssymb}

\usepackage{subfigure}
\usepackage{subcaption}
\makeatletter
\newcommand\figcaption{\def\@captype{figure}\caption}
\newcommand\tabcaption{\def\@captype{table}\caption}
\makeatother
\title{Model Evolution Framework with Genetic Algorithm for Multi-Task Reinforcement Learning}

\author{
Yan Yu$^1$
\and
Wengang Zhou$^1$\and
Yaodong Yang$2$\and
Wanxuan Lu$^3$\and
Yingyan Hou$^3$\and
Houqiang Li$^1$\\
\affiliations
$^1$University of Science and Technology of China\\
$^2$Institute for AI, Peking University\\
$^3$Aerospace Information Research Institute\\
\emails
1140730050@qq.com,
zhwg@ustc.edu.com,
yaodong.yang@pku.edu.com\\
luwx@aircas.ac.cn
houyy@aircas.ac.cn
lihq@ustc.edu.cn
}
\begin{document}

\maketitle

\begin{abstract}

Multi-task reinforcement learning employs a single policy to complete various tasks, aiming to develop an agent with generalizability across different scenarios.
Given the shared characteristics of tasks, the agent's learning efficiency can be enhanced through parameter sharing.
Existing approaches typically use a routing network to generate specific routes for each task and reconstruct a set of modules into diverse models to complete multiple tasks simultaneously.
However, due to the inherent difference between tasks, it is crucial to allocate resources based on task difficulty, which is constrained by the model's structure.
To this end, we propose a \textbf{M}odel \textbf{E}volution framework with \textbf{G}enetic \textbf{A}lgorithm (MEGA), which enables the model to evolve during training according to the difficulty of the tasks.
When the current model is insufficient for certain tasks, the framework will automatically incorporate additional modules, enhancing the model's capabilities.
Moreover, to adapt to our model evolution framework, we introduce a genotype module-level model, using binary sequences as genotype policies for model reconstruction, while leveraging a non-gradient genetic algorithm to optimize these genotype policies.
Unlike routing networks with fixed output dimensions, our approach allows for the dynamic adjustment of the genotype policy length, enabling it to accommodate models with a varying number of modules.
We conducted experiments on various robotics manipulation tasks in the Meta-World benchmark.
Our state-of-the-art performance demonstrated the effectiveness of the MEGA framework.
We will release our source code to the public. 

\end{abstract}

\section{Introduction}

\begin{figure}[htb]
    \centering
    \includegraphics[width=0.995\columnwidth]{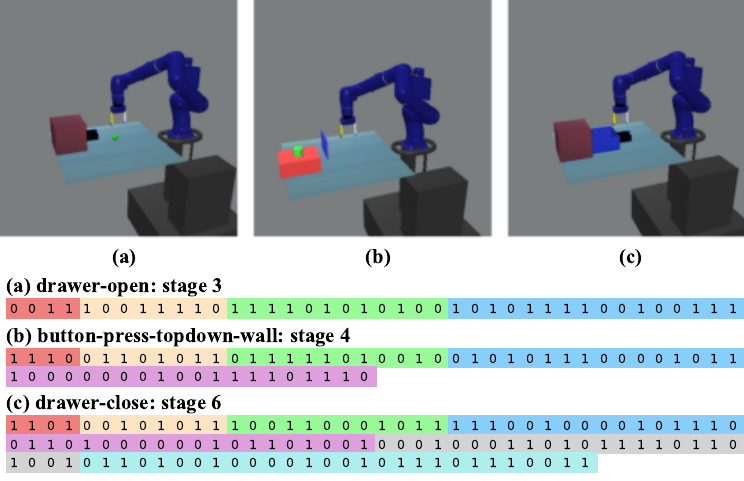}
    \caption{For tasks with varying difficulty, our MEGA uses binary genotype policies with different lengths to allocate varying quantities of modules. The `stage' denotes the number of modules allocated to the task. A segment of the genotype policy with a length of $4$ is used to generate a single module weight. Different segments of the genotype policy, represented by distinct colors, generate module weights at different levels. For a task at stage $N$, the final segment of the genotype policy generates $N+1$ weights, which are used to weight both the initial input and the outputs of $N$ modules.}
    \label{fig:task_gene}
\end{figure}

Reinforcement learning (RL) has made significant progress in recent years, achieving remarkable successes across various domains, including games \cite{berner2019dota,Vinyals}, robotics \cite{kalashnikov2018scalable} and autonomous driving \cite{Kiran_Sobh_Talpaert_Mannion_Sallab_Yogamani_Pérez_2020}. Although RL algorithms have demonstrated impressive capabilities in single-task scenarios, they still encounter challenges when it comes to generalizing across multiple tasks using a single policy.

In previous work, a common approach is to decompose a model into multiple modules \cite{Yang_Xu_Wu_Wang_2020}, which are then reconstructed into different models for each task. These methods typically train a routing network to determine how modules are selected and combined for specific tasks. However, task difficulty varies. Drawing from human learning experiences, it is essential to allocate more resources to more difficult tasks. 
While existing methods do consider the allocation of modules, they only modify the set of modules selected for a given task within a fixed model. These approaches neither alter the model structure nor explicitly account for task difficulty when allocating modules.

To address the above issues, we propose a model evolution framework. 
Our framework enables dynamic evolution through interactions with the environment, inspired by biological evolution. 
When a task cannot be completed with the current model, the agent evolves additional modules to meet the task's requirements. This dynamic resource allocation not only enhances the agent's ability to tackle complex tasks but also reduces the computational cost for simple tasks. 
Since there is no need to predefine the total number of modules at initialization, the agent is not constrained by a lack of sufficient modules for complex tasks, nor is it affected by degradation or gradient explosion due to excessive network depth, offering both adaptability and robustness.

To generate module weights, routing networks are popularly adopted. 
However, routing networks can only generate fixed-dimensional module weights, which are unable to adapt to dynamic model structures.
To this end, we introduce the genotype module-level model. In this model, module weights for different tasks are determined by binary genotype policies, as shown in Fig. \ref{fig:task_gene}. The length of the binary genotype policy is flexible, enabling the generation of module weights with various dimensions, which can adapt to the model evolution framework as modules increase. We categorize genotype policies according to their corresponding tasks into task populations, which collectively form a multi-task community. We optimize these task populations using a non-gradient genetic algorithm, applying operators such as evaluation, selection, crossover, and mutation, enabling the module weights to evolve towards specialization for their respective tasks.

To validate the effectiveness of our framework, we conducted extensive experiments on the Meta-World benchmark \cite{Yu_Quillen_He_Julian_Hausman_Finn_Levine_2019}. 
It can handle 10 to 50 robotic manipulation tasks simultaneously, achieving state-of-the-art performance and sample efficiency compared to other baselines. 
We also made ablation studies, which further demonstrate the significance of our proposed dynamic network structures for multi-task reinforcement learning.

\section{Related Work}
\subsection{Evolutionary Reinforcement Learning}
Evolutionary Algorithms (EA) \cite{Bäck_Schwefel_1993} provide a gradient-free optimization approach.
Evolutionary Reinforcement Learning (ERL) \cite{Pourchot_Sigaud_2018,Khadka_Majumdar_Nassar_Dwiel_Tumer_Miret_Liu_Tumer_2019,chen2019restart,Bodnar_Day_Lió_2020,Bodnar_Day_Lió_2020} combines RL agents with a genetic population, improving the exploration and robustness of the policy. 
For instance, data collected by the EA population is used to train the RL agent~\cite{Khadka_Tumer_2018}, and the RL agent is periodically integrated into the EA population to improve performance. 
A method for shared nonlinear state representations \cite{Li_Tang_Hao_Zheng_Fu_Meng_2022} was introduced, where evolutionary algorithms operate only on the linear policy representations.
However, these methods primarily apply evolution to the parameter networks rather than the model structure itself. 
NEAT \cite{Stanley_Miikkulainen_2002} applies evolutionary algorithms to the topology of neural networks, optimizing the network structure to improve agent performance.  
Despite its potential, these methods often lead to excessively large model structures during training, making it challenging to handle complex tasks. 

HyperNEAT \cite{Stanley_Clune_Lehman_Miikkulainen_2018} further extends neuroevolution to the neural networks building blocks, hyperparameters, architectures, and even algorithms themselves.
PathNet \cite{Fernando_Banarse_Blundell_Zwols_Ha_Rusu_Pritzel_Wierstra_2017}, applied to continual learning problems, uses the genetic algorithm to search for optimal module paths in a model with multiple layers and modules. Once a task is trained, the optimal path and related modules for that task are fixed, and PathNet continues training for the next task.
However, it requires a large number of modules and incurs high training costs. 
Our method allows for the simultaneous training of multiple tasks, improving learning efficiency and reducing the need for excessive modules by sharing parameters.

\subsection{Multi-Task Reinforcement Learning}
Multi-task learning \cite{Caruana_Pratt_Thrun_2017,yang2017multi} aims to improve the efficiency of learning multiple tasks simultaneously by leveraging the similarities among them. 
However, when sharing parameters across tasks, gradients from different tasks may conflict. 
Policy distillation \cite{Rusu_Colmenarejo_Gulcehre_Desjardins_Kirkpatrick_Pascanu_Mnih_Kavukcuoglu_Hadsell_2015,Teh_Bapst_Czarnecki_Quan_Kirkpatrick_Hadsell_Heess_Pascanu_2017,Parisotto_Ba_Salakhutdinov_2015} mitigates the above issue, but it requires introducing additional networks and policies. 
Some methods~\cite{calandriello2014sparse,xu2020knowledge,liu2021conflict} formulate it as a multi-objective optimization problem, but suffer high computational and memory costs.
The Mixture of Experts (MOE) model~\cite{Eigen_Ranzato_Sutskever_2013} constructs multiple networks as experts to improve generalization and reduce overfitting.
The Mixture of Orthogonal Experts (MOORE)
\cite{Hendawy_Peters_D’Eramo_2023} uses a Gram-Schmidt process to shape the shared subspace of representations generated by expert mixtures, encapsulating common structures between tasks to promote diversity.
The Contrastive Module with Temporal Attention (CMTA) \cite{Lan_Zhang_Yi_Guo_Peng_Gao_Wu_Chen_Du_Hu_et} method constrains modules to be distinct from each other and combines shared modules with temporal attention, improving the generalization and performance across multiple tasks.

To further improve the efficiency of parameter sharing, a layer-level model \cite{Yang_Xu_Wu_Wang_2020} has been proposed. It uses a routing network to generate module weights for each task, allowing a set of modules to be reconstructed into task-specific models.
To address the issue of resource allocation for tasks of varying difficulty, Dynamic Depth Routing (D2R) \cite{He_Li_Zang_Fu_Fu_Xing_Cheng_2023} combined a routing network with the module-level model to dynamically allocate modules for different tasks. 
In contrast, our approach explicitly integrates task difficulty with the model structure, enabling the model to evolve adaptively throughout the training process.

\section{Preliminaries}
\begin{figure*}[t!]
    \centering
    \includegraphics[width=0.8\textwidth]{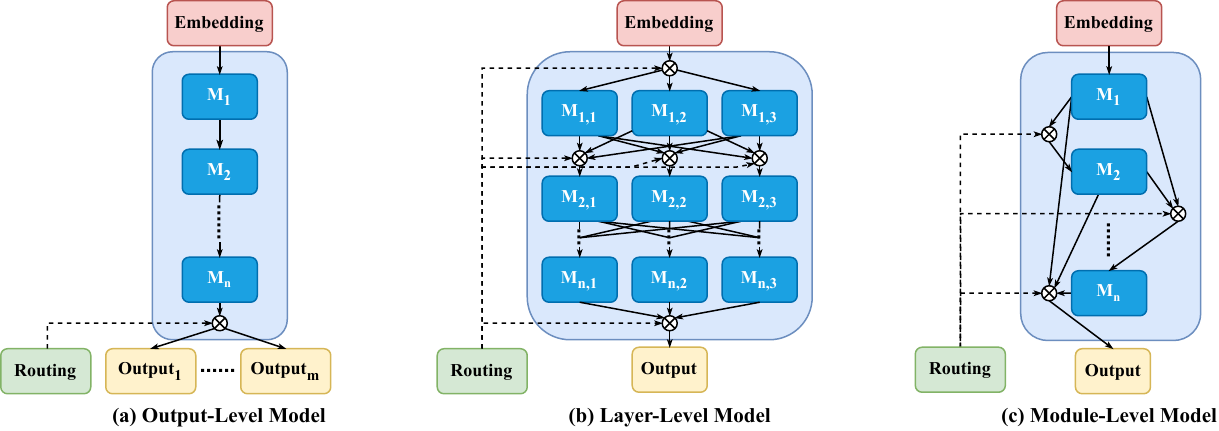}
    \caption{Models with different structures. (a) The model handles each task with a specific output head. (b) The routing network generates module weights for model reconstruction. (c) The routing network selects specific module combinations for different tasks.}
    \label{fig:models}
\end{figure*}
\subsection{Problem Formulation}
The problem studied in this work can be defined as a set of Markov Decision Processes (MDPs) that need to be solved simultaneously. Each MDP is represented as a 5-tuple $\mathcal{M} = \left\langle \mathcal{S}, \mathcal{A}, P, R, \gamma \right\rangle$, where $\mathcal{S}$ is the state space and $\mathcal{A}$ is the action space. 
The dimensions of both the state space and action space are identical for each task and typically share the same meaning in the real world, which forms the foundation for knowledge sharing in Multi-Task Reinforcement Learning (MTRL). 
$P$ is the state transition function $P: \mathcal{S} \times \mathcal{A} \rightarrow \mathcal{S}$. $R$ is the reward function $R: \mathcal{S} \times \mathcal{A} \rightarrow \mathbb{R}$. The reward function is crucial for distinguishing the objectives of each task for the agent. $\gamma \in [0,1)$ is the discount factor. In multi-task reinforcement learning, our goal is to enable the agent to maximize the average 
accumulative reward across all tasks, thereby enhancing the agent's ability to generalize across tasks.

\subsection{Soft Actor-Critic}
We train our framework with the Soft Actor-Critic (SAC) \cite{haarnoja2018soft} algorithm. It introduces soft policy update and maximum entropy during optimization, improving exploration and efficiency. SAC uses a soft Q-function $Q_{\theta}(s_t, a_t)$ as the critic network parameterized by $\theta$, and an actor network $\pi_{\phi}(a_t|s_t)$ parameterized by $\phi$. Besides, SAC introduces a learnable temperature parameter $\alpha$ as a penalty for entropy. The optimization objective is defined as:

\begin{equation}
    \mathcal{J}_{\pi}(\phi) = \mathbb{E}_{s_t \sim \mathcal{D}}\left[\mathbb{E}_{a_t \sim \pi_{\phi}}\left[\alpha \log \pi_{\phi}(a_t | s_t) - Q_{\theta}(s_t, a_t)\right]\right],
\end{equation}
where $\mathcal{D}$ is the data sampled from the replay buffer.
To maintain the entropy of the policy, SAC also introduces the following objective to optimize $\alpha$:

\begin{equation}
\label{eq:alpha}
\mathcal{J}(\alpha) = \mathbb{E}_{a_t \sim \pi_{\phi}}\left[-\alpha \log \pi_{\phi}(a_t | s_t) - \alpha \bar{\mathcal{H}}\right],
\end{equation}
where $\bar{\mathcal{H}}$ represents the desired minimal entropy. During the optimization process, if $\log \pi_{\phi}(a_t | s_t)$ increases with a decrease in $\bar{\mathcal{H}}$, Eq. \ref{eq:alpha} will control $\alpha$ to increase accordingly.

\subsection{Modular Network} 
To address the challenge of resolving multiple tasks with a single model, several approaches with varying structures have been proposed, as illustrated in Fig. \ref{fig:models}.
The output-level model decomposes the output layer into a multi-head module while sharing the backbone. 
The layer-level model \cite{Yang_Xu_Wu_Wang_2020} consists of multiple layers, each containing several modules. The input to the next layer is the weighted sum of the outputs from the previous layer, with the weights generated by a routing network. 
However, this approach requires that all modules participate in the prediction for each task and does not allow for cross-layer connections. 
The module-level model \cite{He_Li_Zang_Fu_Fu_Xing_Cheng_2023} allows the input to the next module to be a weighted sum of the outputs from a selected subset of previous modules, thus enabling the skipping of several modules.
While this structure facilitates task-specific customization, it retains a fixed architecture throughout training and does not explicitly account for task difficulty. 
Furthermore, the model’s structure is dependent on the predetermined hyperparameter for the number of modules, and both an excessively large or small number of modules can negatively impact the training process.
Our work optimizes the model by integrating the module-level network with the model evolution framework, aiming to address the issues of the fixed model structure and the reliance on hyperparameters.

\section{Method}
In this section, we first introduce the overall structure of the MEGA framework and elaborate its core components. Then, we explain how the genotype policy is used to reconstruct the model, and how genetic algorithms are applied to select, crossover, and mutate the genotype policies within the task population. 
Finally, we introduce the model evolution framework, which enables continuous evolution in response to increasingly complex tasks, dynamically allocating different amounts of resources to each task based on its difficulty.

\subsection{Genotype Module-Level Model}
\begin{figure}
    \centering
    \includegraphics[width=0.95\columnwidth]{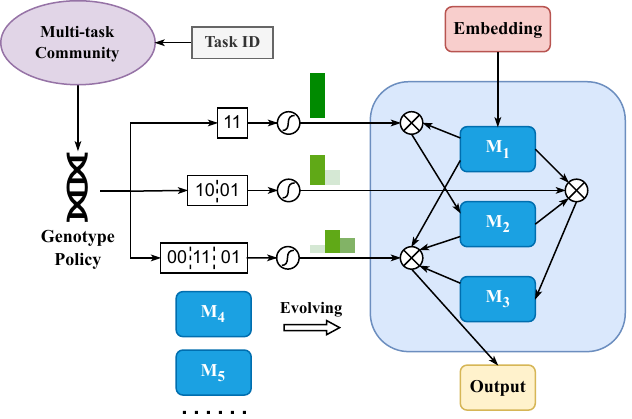}
    \caption{The structure of the genotype module-level model involves selecting a genotype policy from the multi-task community based on the task ID, decomposing the genotypes into weights, and reconstructing the module-level model. To enhance the model's capabilities, the model continuously evolves by incorporating additional modules.}
    \label{fig:Overll}
\end{figure}

The overall structure of the genotype module-level model is shown in Fig. \ref{fig:Overll}, which includes the genotype policy selection and the module-level model. When the agent handles a task, it first selects a genotype policy from the multi-task community based on the task ID. The genotype is then split and transformed into module weights. 
After embedding, the states are converted into the initial output \( m_0 \), which is then passed as input to the first module. The input to each subsequent module is the weighted sum of the outputs from the previous modules as follows:
\begin{gather}
m_0 = \text{embedding}(s), \\
m_i = M_i \left( \sum_{j=0}^{i-1} w_{i,j} \cdot m_j \right).
\end{gather}
The output dimension of each layer is consistent and can be fed into the execution module to generate the action.

\subsection{Genotype Policy}
The genotype policy determines the weights of the model's output and is represented by a binary sequence, which facilitates optimization through genetic algorithms. The length of the genotype policy is determined by the precision of the weights \( p_w \) and the number of modules \( n_m \). For the \( i \)-th level module, its input consists of the weighted sum of the state representation module and the outputs of the previous \( i-1 \) modules, requiring \( i \) weights. Therefore, the total length of the genotype policy is given by:
\begin{equation}
    l_g = \frac{1}{2} p_w (n_m + 1) n_m.
\end{equation}

When converting the genotype policy into module weights, we first decompose the genotype policy according to the number of layers and modules and then convert the corresponding binary sequences into decimal values \( d_{i,j} \), where each \( d_{i,j} \) represents the weight corresponding to the $m_j$ when it is input to the \( i \)-th module.
Subsequently, the decimal values \( d_{i,j} \) using the Softmax function to normalize them into weights. It is important to highlight that the binary-to-decimal conversion yields values within the range \( [0, 2^{p_w} - 1] \). After applying the Softmax function, the weight values for each input to the \( i \)-th module are constrained to the interval:
\begin{equation}
w_{i,j}\in\left[\frac{1}{1+(i-1)e^{2^{p_w}-1}}, \frac{e^{2^{p_w}-1}}{e^{2^{p_w}-1}+i-1}\right].
\end{equation}
The weights obtained from the Softmax function do not effectively cover the full range of \( [0, 1] \) reduces the model's ability to select the optimal module for inference. Further details can be found in the supplementary material. To address this issue, we apply the HalfSoftmax function to convert the genotype policy into weights:
\begin{equation}
\begin{aligned}
w_{i,j} &= \text{HalfSoftmax}(d_{i,0},\cdots, d_{i,i-1})\\
&=\frac{e^{d_{i,j}} - 0.99}{\sum_{n=0}^{i-1} e^{d_{i,n}} - 0.99 \cdot i} .
\label{Eq:Proposition-1}
\end{aligned}
\end{equation}
When \( d_{i,j} = 0 \), the corresponding weight becomes extremely small, which helps reduce the influence of irrelevant modules on the final output.

\subsection{Genetic Algorithm}
The multi-task community contains all genotype policies, each assigned to a specific task population based on the corresponding task. Within each task population, each genotype policy is dedicated to optimizing the same task. We employ the genetic algorithm to optimize each task population, which involves  evaluation, selection, crossover, and mutation operations. 

\begin{algorithm}[tb]
    \caption{Genetic Algorithm}
    \label{alg:Genetic}
    \textbf{Input}: the task number $N_T$, the population size $N_{P}$, the module weight precision $p_{w}$, the minimum module number $n_{m}$, the abandon rate $r_a$.\\
    \textbf{Initialization}: the multi-task community composed of task populations $\mathcal{C}=\{\mathcal{P}_1,\cdots,\mathcal{P}_{N_T}\}$, each task population composed of genotype policies $\mathcal{P}=\{\mathcal{G}_1,\cdots,\mathcal{G}_{N_P}\}$, for each genotype policy $\mathcal{G}$, randomly initialize the binary sequence based on $p_w$ and $n_m$.
    \begin{algorithmic}[1] %[1] enables line numbers
        \STATE Evaluate fitness $f$ each genotype policy $\mathcal{G}$;
        \FOR{$\mathcal{P}$ in $\mathcal{C}$}
        \STATE Check the best genotype policy $\mathcal{G}_{best}$;
        \STATE Abandon the worst $p_w*r_a$ genotype policies;
        \WHILE{$i < \frac{1}{2}p_w*r_a$}
        \STATE $\mathcal{G}_{p_1}, \mathcal{G}_{p_2} \leftarrow \text{Choose\_Crossover\_Parents}(\mathcal{P})$;
        \STATE $\mathcal{G}_{c} \leftarrow \text{Crossover}(\mathcal{G}_{p_1}, \mathcal{G}_{p_2} )$;
        \STATE $\mathcal{G}_{p_m} \leftarrow \text{Choose\_Mutate\_Parent}(\mathcal{P})$;
        \STATE $\mathcal{G}_{m} \leftarrow \text{Mutate}(\mathcal{G}_{p_m})$;
        \STATE Insert $\mathcal{G}_{c}, \mathcal{G}_{m}$ to $\mathcal{P}$;
        \STATE $i\leftarrow i+1$;
        \ENDWHILE
        \ENDFOR
        \RETURN $\mathcal{C}$.
    \end{algorithmic}
\end{algorithm}
In multi-task reinforcement learning, evaluating each genotype policy for every task can be computationally expensive. To address this, we perform evaluations concurrently during training. When starting a task, we first select a genotype policy from the corresponding task population. Unevaluated genotype policies are given priority for selection. If all genotype policies have been evaluated, a policy is randomly selected from the task population. The currently optimal genotype policy has a higher probability of being selected to ensure the agent samples higher-quality data. This data sampling strategy also increases the entropy of the sampled policies, promoting better exploration. After completing a simulation, the selected genotype policy receives its total reward as its fitness, which serves as a key reference for determining whether the genotype policy is retained and undergoes reproduction.

\begin{figure}[htb]
    \centering
    \includegraphics[width=0.95\columnwidth]{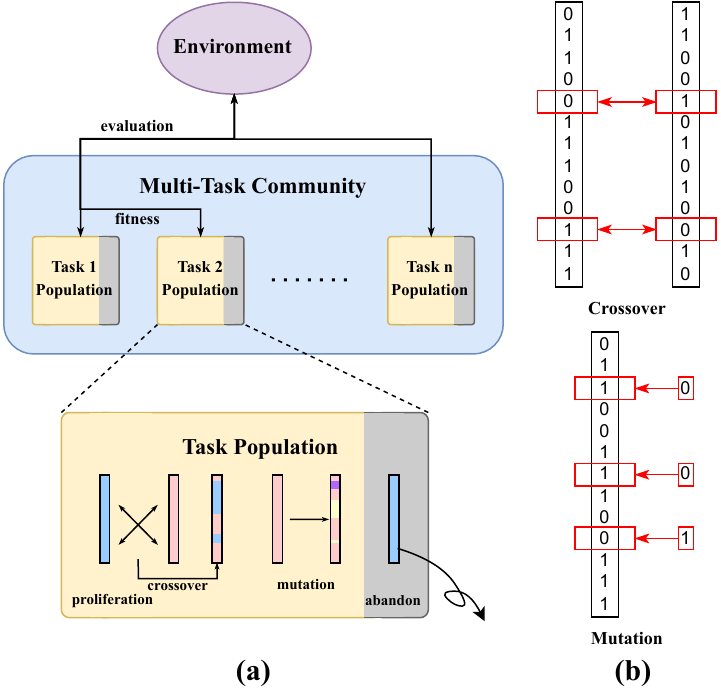}
    \caption{The genetic algorithm optimizes the genotype policy population as follows. (a) For each task, a genotype policy is selected from the task population to perform the task, and its reward serves as fitness. During optimization, crossover and mutation operations are applied within the task population, while less fit policies are discarded. (b) The crossover operator exchanges parts of two genotype policies to create new ones, while the mutation operator modifies a single genotype policy by flipping some of its genes.}
    \label{fig:evoluation}
\end{figure}

After evaluation, each task population will discard a certain percentage of genotype policies based on their fitness. The remaining genotype policies are then optimized using crossover and mutation operators. The crossover operator selects two genotype policies from a task population and exchanges parts of their genes to form a new genotype policy. The mutation operator selects one genotype policy and flips part of its genes to generate a new genotype policy.
The optimal genotype policies are given a higher probability of reproduction, allowing the task populations to evolve continuously in an optimal direction while maintaining diversity. The details of the genetic algorithm are shown in Fig. \ref{fig:evoluation} and Alg. \ref{alg:Genetic}. The pseudo codes for the crossover and mutation operators are provided in supplementary material.
Due to the lack of a suitable evaluation metric for the critic network, we continue using the module-level model as the critic.

\subsection{Model Evolution}

\begin{figure}
    \centering
    \includegraphics[width=0.5\columnwidth]{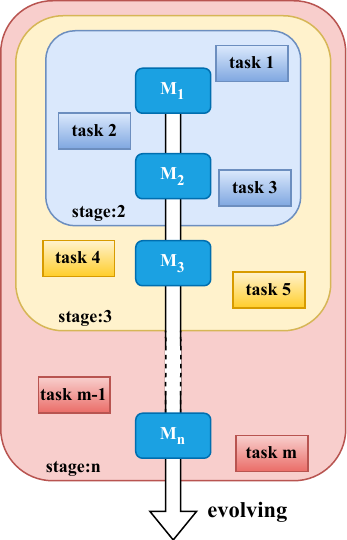}
    \caption{The mechanism of the model evolution framework. When tasks cannot be completed by the current model, it dynamically adds modules and evolves to adapt to the more difficult tasks.}
    \label{fig:Evoluationary_model}
\end{figure}

To address the issue of the predetermined number of modules, we propose the model evolution framework, as shown in Fig. \ref{fig:Evoluationary_model}. Each task is assigned a stage, which determines the maximum number of modules the agent should use for that task. If the agent is unable to complete the task, the stage gradually increases during the training process until the agent can either complete the task or make significant progress on it. This design ensures that, in the early stages of training, the agent uses a minimal module quantity, enabling efficient learning for easier tasks or skills. It also enables the agent to allocate more resources focus on difficult tasks, as additional modules will not be used to handle the tasks that are already solved. Ideally, this approach enables the agent to train specialized modules for common sub-skills, facilitating knowledge sharing and reuse across multiple tasks.

The length of a genotype policy is determined by the stage of the corresponding task. When the stage of a task increases due to it is unable to be completed, we discard all genotype policies whose length does not meet the current stage during the optimization of the task population. These discarded policies are replaced by randomly generated genotype policies that meet the required length. As the task populations are independent, the replacement of a genotype policy for one task does not affect the training of other tasks, ensuring the stable training of the model evolution framework.

\begin{figure*}[hbtp]
    \centering
    \includegraphics[width=\textwidth]{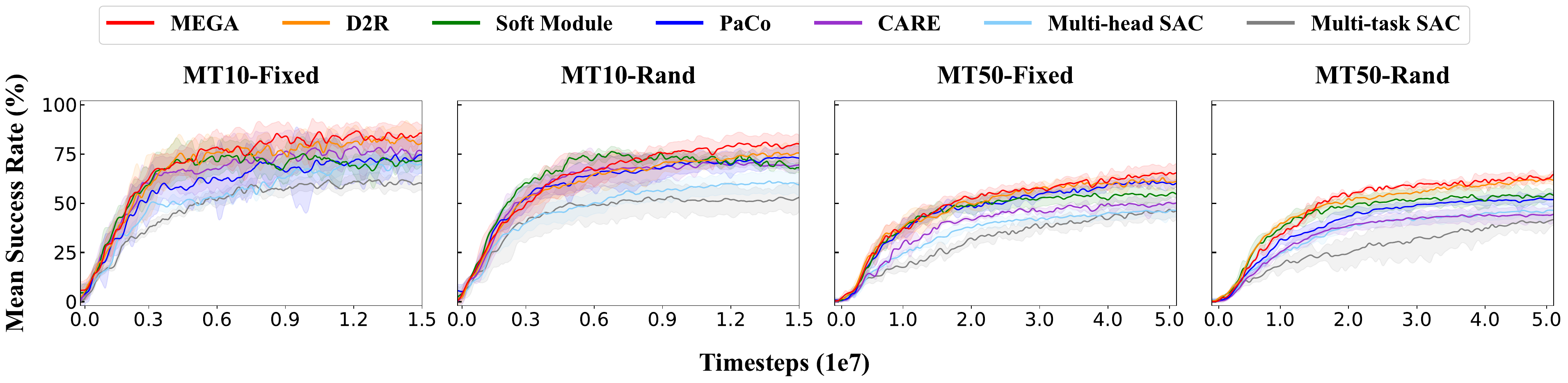}
    \caption{Comparison of our method against baselines in Meta-World benchmark with 10 \& 50 tasks and fixed \& rand  goal setting.}
    \label{fig:result}
\end{figure*}

\section{Experiments}
In this section, we conduct several experiments to investigate the following questions:
\begin{itemize}
    \item Whether the MEGA framework contribute to improved performance and learning efficiency?
    \item Is the module-level model sensitive to the predetermined number of modules, and can the model evolution framework address this issue?
    \item Does the genetic algorithm effectively optimize the task populations to specialize in specific tasks?
\end{itemize}

\begin{table}[htbp]
    \tabcaption{Comparison of the average success rates and variances between and the baseline algorithms and our MEGA across four benchmark settings.}
    \centering
    \tiny
    \begin{tabular}{c|cccc}
        \toprule
        Algorithms& \textbf{MT10-Fixed} & \textbf{MT10-Rand} &  \textbf{MT50-Fixed} &  \textbf{MT50-Rand} \\
        \midrule
        D2R          & $0.86\pm 0.05$ & $0.77\pm 0.05$ &  $0.64\pm 0.03$ &  $0.63\pm 0.04$\\
        Soft Module  & $0.76\pm 0.08$ & $0.78\pm 0.03$ &  $0.56\pm 0.02$ &  $0.55\pm 0.03$\\
        PaCo         & $0.76\pm 0.05$ & $0.74\pm 0.05$ &  $0.62\pm 0.03$ &  $0.53\pm 0.02$\\
        CARE         & $0.79\pm 0.08$ & $0.71\pm 0.02$ &  $0.50\pm 0.04$ &  $0.50\pm 0.04$\\
        MTMH-SAC     & $0.71\pm 0.05$ & $0.62\pm 0.05$ &  $0.46\pm 0.04$ &  $0.46\pm 0.03$\\
        MT-SAC       & $0.63\pm 0.05$ & $0.55\pm 0.07$ &  $0.47\pm 0.04$ &  $0.42\pm 0.04$\\
        \midrule
        MEGA          & $\textbf{0.88}\pm \textbf{0.04}$ & $\textbf{0.82}\pm \textbf{0.05}$ &  $\textbf{0.66}\pm \textbf{0.05}$ &  $\textbf{0.65}\pm \textbf{0.02}$\\
        \bottomrule
    \end{tabular}
    \label{tab:result}
\end{table}
\subsection{Environment Setup}
We chose the Meta-World environment as our benchmark to evaluate the MEGA framework.
Meta-World provides 50 different robotics continuous control and manipulation tasks, making it an effective benchmark for evaluating a model's generalization capability in multi-task settings. 
The environment includes two main task settings, \textbf{MT10} and \textbf{MT50}, which require the agent to simultaneously learn 10 and 50 tasks.
Additionally, all tasks are extended to a random-goal setting, further increasing the challenge of the benchmark. Consequently, our experimental results include both fixed goal settings, \textbf{MT10-Fixed} and \textbf{MT50-Fixed}, as well as random goal settings, \textbf{MT10-Rand} and \textbf{MT50-Rand}.

\subsection{Baseline Algorithms}
We chose SAC to train our framework, as it has been proven effective in the Meta-World environment \cite{Yu_Quillen_He_Julian_Hausman_Finn_Levine_2019}.
The MEGA framework is built upon the SAC \cite{haarnoja2018soft} algorithm. To evaluate our approach, we compare it against six SAC-based baseline algorithms:
(a) \textbf{Multi-task SAC (MT-SAC)}: This approach combines a one-hot task ID with the observation as the model input.
(b) \textbf{Multi-task Multi-head SAC (MTMH-SAC)}: An enhancement of MT-SAC based on the output-level model. It uses a shared backbone network to process features and  multiple heads to handle each task.
(c) \textbf{Soft-Module} \cite{Yang_Xu_Wu_Wang_2020}: The layer-level model that employs a routing network to generate module weights for model reconstruction.
(d) \textbf{Dynamic Depth-Routing (D2R)} \cite{He_Li_Zang_Fu_Fu_Xing_Cheng_2023}: The module-level model that selects a different module quantity for each task.
(e) \textbf{CARE} \cite{Sodhani_Zhang_Pineau_2021}: Utilizing a mixture of encoders to learn fine-grained control over information across tasks. 
(f) \textbf{PaCo} \cite{Sun_Zhang_Xu_Tomizuka_2022}: Learning representative parameter sets used to compose policies for different tasks.
\subsection{Quantitative Results}
We trained our MEGA framework and baseline algorithms on MT10-Fixed, MT10-Rand, MT50-Fixed and MT50-Rand. 
The performance of each algorithm was evaluated based on its task success rate. 
Each algorithm was run across 5 seeds, with training conducted for 7,500 episodes per task in MT10 and 5,000 episodes per task in MT50, using 200 timesteps per episode. The results, averaged over the 5 seeds, are shown in Fig. \ref{fig:result} and Tab. \ref{tab:result}.
Compared to the baseline algorithms, the MEGA framework consistently achieves the highest success rates in both fixed and random goal settings for the MT10 and MT50 tasks. 
We also compared the module utilization efficiency between MEGA and D2R, which share a similar structure. 
As shown in Fig. 2, the MEGA model uses fewer modules for each task compared to D2R, with an average reduction of $5.2$ modules per task, which further validates the effectiveness and efficiency of MEGA.

\begin{figure}[h]
    \centering
    \includegraphics[width=0.9\columnwidth]{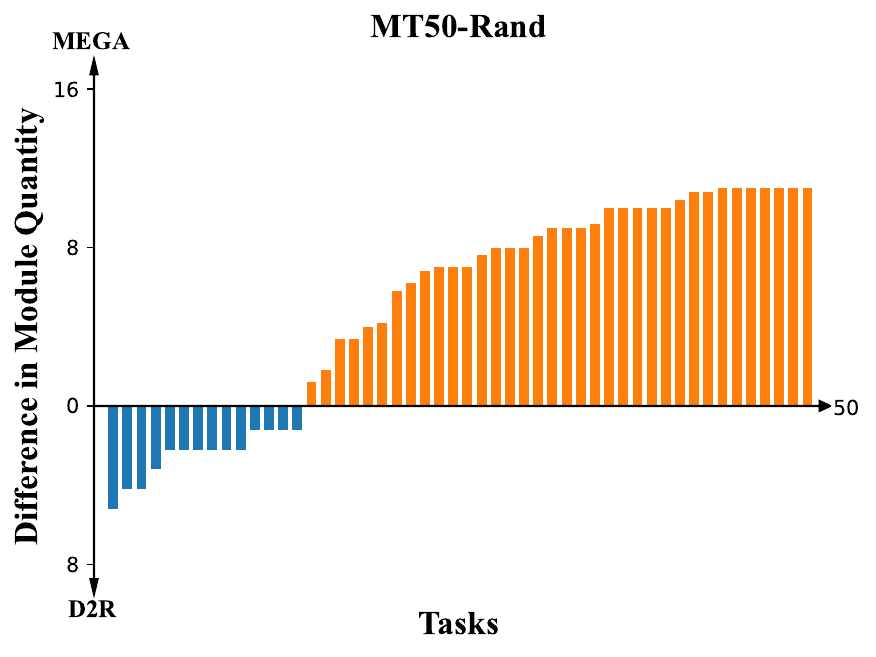}
    \caption{The difference in module quantity between MEGA and D2R in the MT50-Rand setting. The orange bars represent the number of modules MEGA uses fewer than D2R, while the blue bars represent the number of modules D2R uses fewer than MEGA.}
    \label{fig:stages}
\end{figure}

\begin{figure*}[h]
    \centering
    \includegraphics[width=0.98\textwidth]{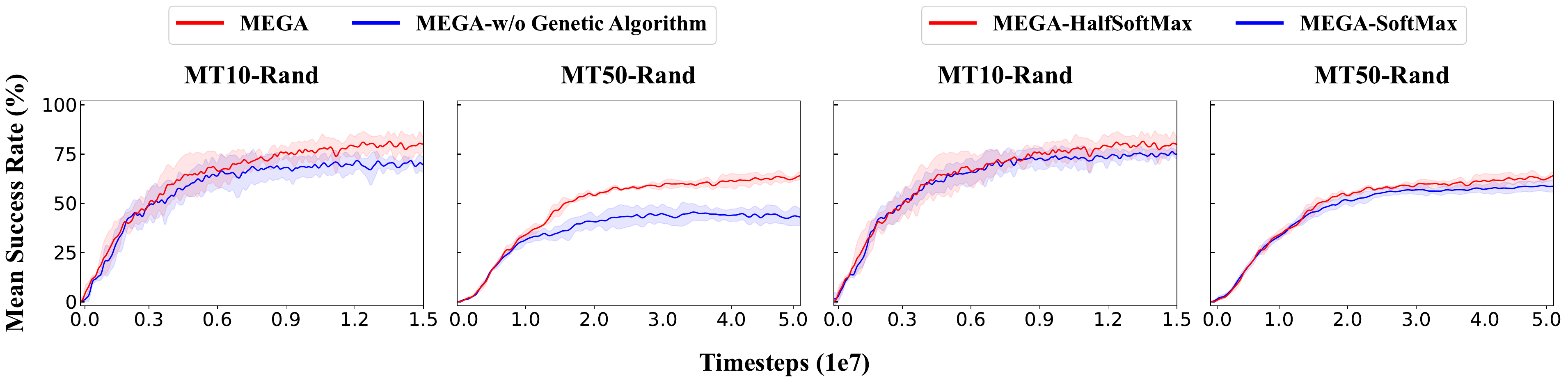}
    \caption{Ablation studies on the genetic algorithm and HalfSoftMax function.}
    \label{fig:ablation2}
\end{figure*}

\begin{figure}[h]
    \centering
    \includegraphics[width=0.77\columnwidth]{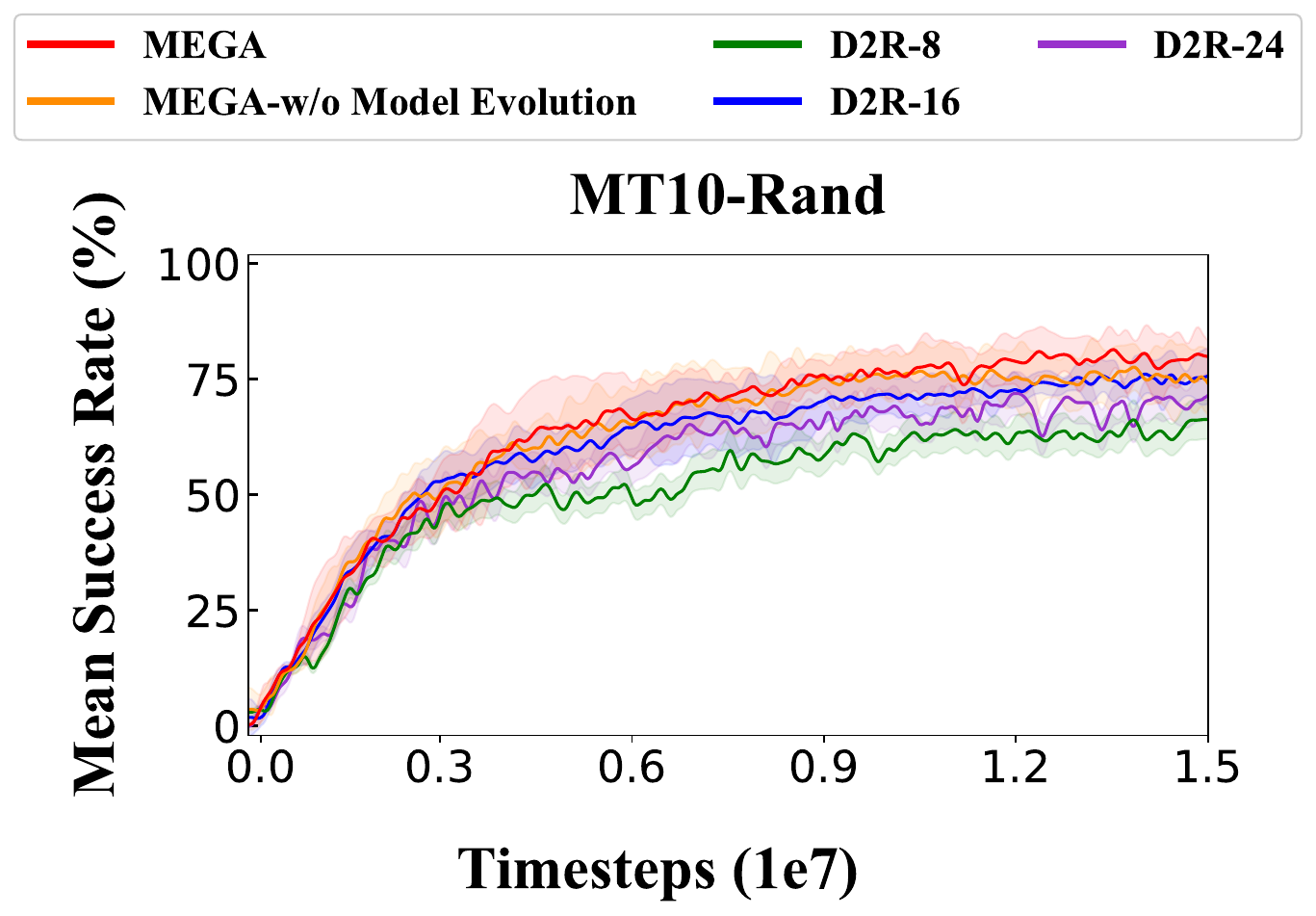}
    \caption{Ablation study on model evolution framework and module quantity.}
    \label{fig:ablation1}
\end{figure}

\subsection{Ablation Study}
\subsubsection{Impact of Model Evolution Framework}
The model without the evolutionary mechanism is configured with $16$ modules, consistent with the D2R setting. Compared to the full MEGA framework, the performance of this model is reduced and closer to that of D2R, as shown in Fig. \ref{fig:ablation1}. 
This result indicates that the performance improvement in MEGA is primarily due to the evolutionary mechanism.Additionally, it suggests that the optimization of the non-gradient evolutionary algorithm is comparable to that of the routing network.
We also evaluated the model performance with different maximum module quantities. 
Models with $8$ and $24$ modules performed worse than the model with 16 modules, demonstrating that both insufficient and excessive module quantity negatively impact performance.
Therefore, the model is sensitive to the module quantity, emphasizing the significance of dynamic model structure.

\begin{figure}
\centering
    \includegraphics[width=0.8\columnwidth]{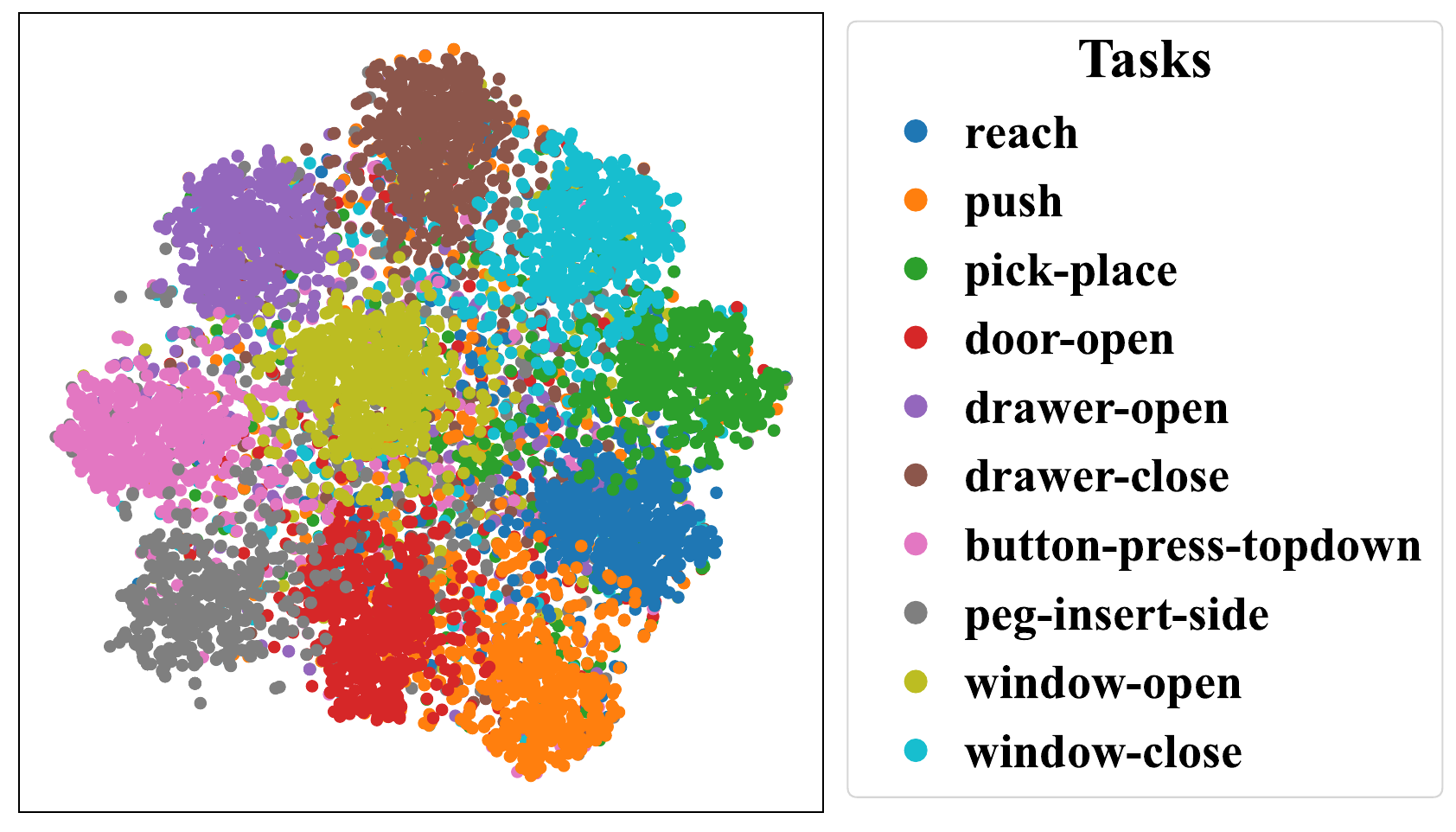}
    \figcaption{t-SNE visualization of the genotype policies across the MT10 tasks.}
    \label{fig:tsne}
\end{figure}

\subsubsection{Impact of Model Genetic Algorithm}
In Fig. \ref{fig:ablation2} (a) and (b), we investigate the impact of the genetic algorithm on model performance. When the genetic algorithm is removed and the model is controlled by randomly generated genotype policies, performance significantly decreases, particularly in the MT50 setting, which demands higher generalization across tasks.
We also present the t-SNE visualization of the genotype policies for each task in Fig. \ref{fig:tsne}.
For each task, we collected multiple sets of genotype policies. Since the genotype policies are generated through crossover and mutation operations, where cross-population breeding may occur, leading to some overlap in the result. Despite this, all the tasks remain clearly distinguishable from one another, indicating that the task populations, composed of genotype policies, are specialized for their respective tasks. 

\subsubsection{Impact of Model HalfSoftmax Function}
We introduced the HalfSoftmax function to convert genotype policies into module weights to enable the model to skip irrelevant modules. In Fig. \ref{fig:ablation2} (c) and (d), we compare the model's performance using the HalfSoftmax and SoftMax functions. The results demonstrate that incorporating HalfSoftmax enhances the model's flexibility, enabling the agent to more efficiently leverage cross-task knowledge, thereby improving both performance and learning efficiency.

\section{Conclusion}
In this work, we propose the MEGA framework, which enables adaptive evolution during the training process in a multi-task setting. 
When the number of modules is insufficient to handle a task, the model evolution mechanism of MEGA can automatically add modules to enhance the model's overall capacity. 
To effectively allocate modules for each task in a dynamic model with varying module quantity, we introduce the genotype module-level model to generate module weights using binary sequence genotype policies. 
By adjusting the length of the genotype policy, the model can adapt to different module quantity. 
e use the genetic algorithm to optimize the genotype policies, with the episode reward serving as the fitness for each task. 
Based on fitness, high-quality genotype policies are selected for generating new genotypes by crossover and mutation operations, guiding the task population to evolve towards task specialization.
Our experiments in the Meta-World benchmark show that the MEGA framework outperforms several baseline models.

In future work, we aim to further explore the integration of genetic algorithms with deep reinforcement learning, combining both gradient-based and non-gradient optimization methods to improve model performance and sample efficiency. This will help develop sustainable learning capabilities for reinforcement learning tasks.

\bibliographystyle{named}
\bibliography{ijcai25}

\clearpage

\appendix
\section{Genotype Policy Decomposition}
\begin{figure}[h]
    \centering
    \includegraphics[width=\columnwidth]{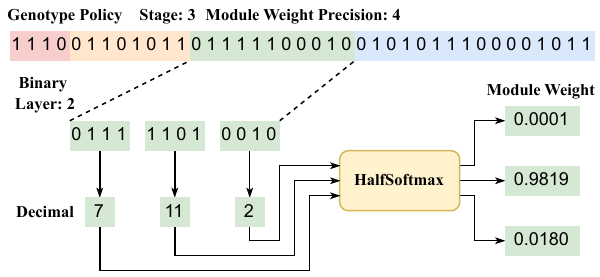}
    \caption{The process of decomposing the genotype policy and converting it into module weights. We selected a genotype policy with a stage of $3$ and a module weight precision of $4$ as an example and transformed the second layer into module weights. The second layer module weight processing involves the embedded state representation \( m_0 \) and the outputs of the first and second modules \( m_1 \) and \( m_2 \), hence it consists of three segments of the genotype.}
    \label{fig:g2w}
\end{figure}

The process of genotype policy decomposition is illustrated in Fig. \ref{fig:g2w}. First, the genotype policy is divided into segments based on the layers, with each segment corresponding to the weights of modules in a single layer. Each segment is then further split according to the module weight precision, where each sub-segment corresponds to the weight of an individual module. Finally, the binary genotype is converted into decimal, and the HalfSoftmax function is applied to transform the decimal values into module weights.

\section{HalfSoftmax Function}

In this section, we describe the calculation method and significance of the HalfSoftmax function.

\begin{figure}
    \centering
    \includegraphics[width=\columnwidth]{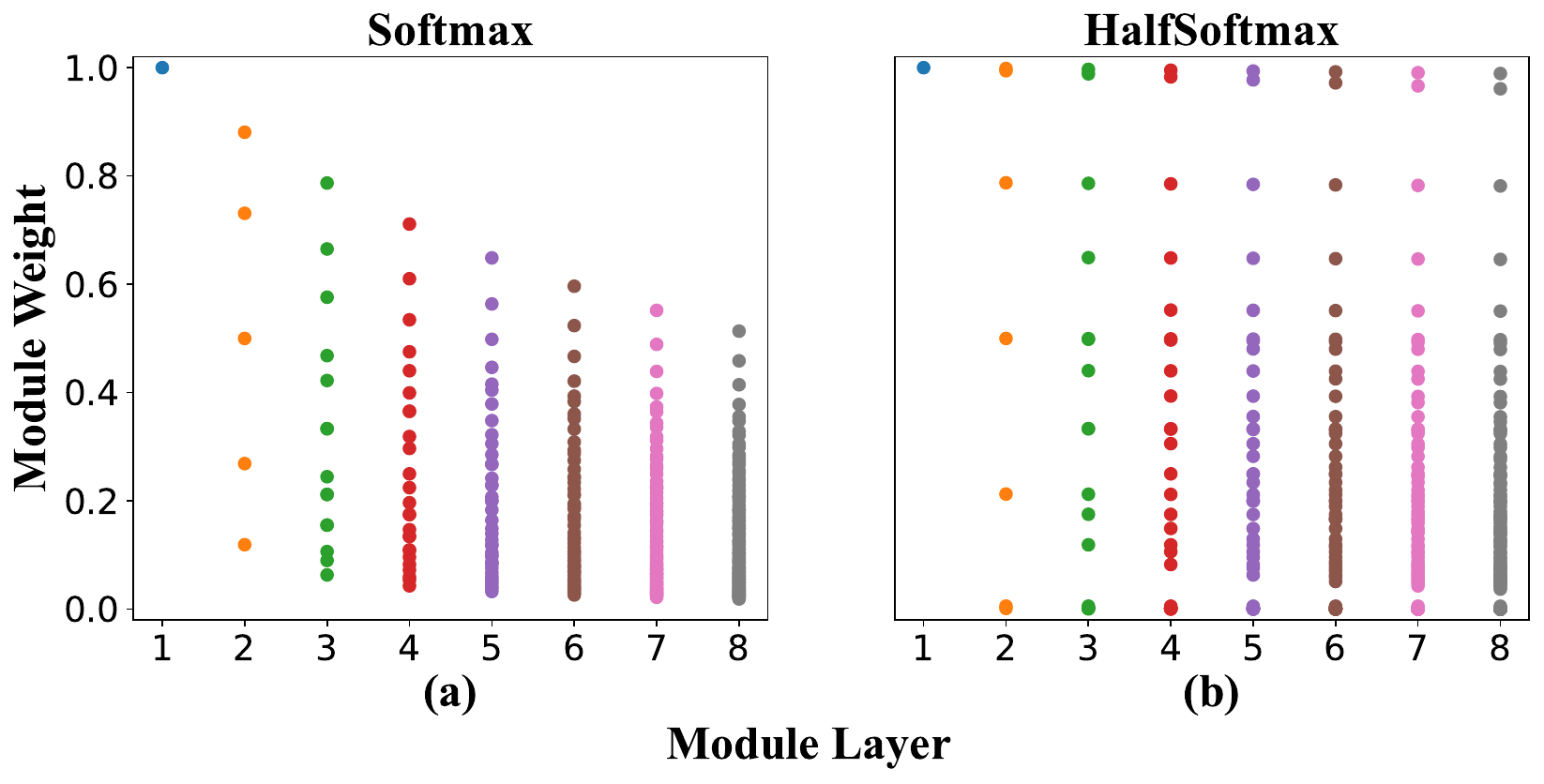}
    \caption{The interval of module weights converted by the Softmax and HalfSoftmax functions with module weight precision of $2$. the lower bound of the module weights generated by the Softmax function for the earlier layers is relatively high, preventing the model from skipping irrelevant modules. As the number of layers increases, the upper bound of the module weights gradually decreases, reducing the coverage range and limiting the ability to emphasize important modules. In contrast, the HalfSoftmax function generates weights that more effectively cover the entire range across all layers, allowing for stronger module expressiveness. Notably since the first layer contains only one module weight, it has only one value of $1$.}
    \label{fig:halfsoftmax}
\end{figure}

Due to the limited precision of the genotype policy, the generated weights are constrained to discrete values, which restricts the range of possible weights. As shown in Fig. \ref{fig:halfsoftmax} (a), the module weights generated by the Softmax function for each layer exhibit this limitation. Specifically, for the earlier layers, the minimum module weight fails to reach 0, preventing the model from skipping irrelevant modules. As the number of layers increases, the maximum weight value decreases, which affects the expression of important modules.

To enhance the genotype policy's ability to select modules, we introduce the HalfSoftmax function. Specifically, after converting the binary values into decimals, an exponential operation is applied, followed by subtracting $0.99$. The resulting value is then normalized to obtain the weight. The formula is as follows:

\begin{gather}
\tilde{d}_{i,j} = e^{d_{i,j}} - 0.99,\\
w_{i,j} = \frac{\tilde{d}_{i,j}}{\sum_{n=0}^{i-1}\tilde{d}_{i,j}} = \frac{e^{d_{i,j}} - 0.99}{\sum_{n=0}^{i-1} e^{d_{i,n}} - 0.99 \cdot i}.
\end{gather}

Fig. \ref{fig:halfsoftmax} (b) shows the range of module weights generated by the HalfSoftmax function for each layer. As observed, the values generated by the HalfSoftmax function more fully cover the interval $[0, 1]$, enabling a more precise selection of relevant modules for model inference.

\section{Pseudo Code}\label{Pseudo Code}
We present the pseudo code for the method used to select parent genotype policies for crossover and mutation in Alg. \ref{alg:Genetic}, along with the pseudo code for the crossover and mutation operators.
\begin{algorithm}[htb]
    \caption{Choose Crossover Parents for Task $i$}
    \label{alg:Choose Crossover}
    \textbf{Input}: the multi-task community $\mathcal{C}$, the task population $\mathcal{P}_i$ for the task $i$, the crossover cross population threshold $\tau_{cp\_c}\in(0,1)$
    \begin{algorithmic}[1] %[1] enables line numbers
        \STATE Init random number $n\in[0,1]$
        \IF {$n > \tau_{cp\_c}$}
        \STATE $\mathcal{G}_1\leftarrow $ the best genotype policy $\mathcal{G}_{\text{best}_i}$ of $\mathcal{P}_i$ 
        \STATE $\mathcal{G}_2\leftarrow \text{Random\_Choose}(\mathcal{C})$ 
        \ELSE
        \STATE $\mathcal{G}_1, \mathcal{G}_2\leftarrow \text{Random\_Choose}(\mathcal{P}_i)$ 
        \ENDIF
        \RETURN $\mathcal{G}_1, \mathcal{G}_2$
    \end{algorithmic}
\end{algorithm}

\clearpage
\begin{algorithm}[htb]
    \caption{Crossover}
    \label{alg:Crossover}
    \textbf{Input}: the parents genotype policies $\mathcal{G}_{1}, \mathcal{G}_2$, the crossover rate $r_c$
    \begin{algorithmic}[1] %[1] enables line numbers
        \STATE $\mathcal{G}_\text{long}, \mathcal{G}_\text{short}\leftarrow \text{Sort}(\mathcal{G}_1, \mathcal{G}_2)$ \\
        // Sort parents genotype policies based on its length
        \STATE $l\leftarrow \text{Length}(\mathcal{G_{\text{short}}})$
        \STATE $n_c\rightarrow l*r_c$
        \STATE $\text{Crossover\_Poses} \leftarrow \text{Random\_Choose\_n}(l, n_c)$\\
        // Choose $n_c$ different positions in $[1, l]$ for crossover
        \STATE $\mathcal{G}_{new} \leftarrow \mathcal{G}_{\text{long}}$
        \FOR{$p$ in $\text{Crossover\_Poses}$}
        \STATE
            $\mathcal{G}_{new}[p]\leftarrow \mathcal{G}_{short}[p]$\\
        \ENDFOR
        \RETURN $\mathcal{G}_{new}$
    \end{algorithmic}
\end{algorithm}

\begin{algorithm}[htb]
    \caption{Choose Mutate Parents for Task $i$}
    \label{alg:Choose Mutate}
    \textbf{Input}: the multi-task community $\mathcal{C}$, the task population $\mathcal{P}_i$ for the task $i$, the mutate cross population threshold $\tau_{cp\_m}\in(0,1)$, the mutate best genotype threshold $\tau_{b\_m}\in(0, \tau_{cp\_m})$
    \begin{algorithmic}[1] %[1] enables line numbers
        \STATE Init random number $n\in[0,1]$
        \IF {$n > \tau_{cp\_cm}$}
        \STATE $\mathcal{G}\leftarrow \text{Random\_Choose}(\mathcal{C})$ 
        \ELSIF{$n > \tau_{b\_cm}$}
        \STATE $\mathcal{G}\leftarrow \text{Random\_Choose}(\mathcal{P}_i)$ 
        \ELSE
        \STATE $\mathcal{G}\leftarrow $ the best genotype policy $\mathcal{G}_{\text{best}_i}$ of $\mathcal{P}_i$ 
        \ENDIF
        \RETURN $\mathcal{G}$
    \end{algorithmic}
\end{algorithm}

\begin{algorithm}[htb]
    \caption{Mutate}
    \label{alg:Mutate}
    \textbf{Input}: the parent genotype policy $\mathcal{G}$, the mutate rate $r_m$
    \begin{algorithmic}[1] %[1] enables line numbers
        \STATE $l\leftarrow \text{Length}(\mathcal{G})$
        \STATE $n_m\rightarrow l*r_m$
        \STATE $\text{Mutate\_Poses} \leftarrow \text{Random\_Choose\_n}(l, n_m)$\\
        \STATE $\mathcal{G}_{new} \leftarrow \mathcal{G}$
        \FOR{$p$ in $\text{Mutate\_Poses}$}
        \STATE
            $\mathcal{G}_{new}[p]\leftarrow \mathcal{G}[p]$\\
        \ENDFOR
        \RETURN $\mathcal{G}_{new}$
    \end{algorithmic}
\end{algorithm}

\section{Parameters}
Here we present the hyperparameter settings of the experiments and genetic algorithm in Tab. \ref{table:general} and Tab. \ref{table:evolutionary}. Our experiments were conducted on GeForce RTX 2080Ti and GeForce RTX 3070Ti GPUs.

\begin{table}[htbp]
    \centering
  \caption{General settings}
    \label{table:general}
  \centering
  \begin{tabular}{ll}
    \toprule
    \textbf{Parameter}      & \textbf{Value}    \\ 
    \midrule
    Embedding Network Size          & $2 \times 400$                 \\
    Module Input\&Output Dimension  & $400$                          \\
    Module Size                     & $128$                          \\
    Gamma                           & $0.99$                         \\
    Actor Learning Rate             & $3\times10^{-4}$               \\
    Critic Learning Rate            & $3\times10^{-4}$               \\
    $\alpha$ Learning Rate          & $1\times10^{-4}$               \\
    Episode Length                  & $200$                          \\
    Batch Size                      & $1280$ (MT10) / $6400$ (MT50)  \\
    Activation Function             & ReLU                           \\
    Reward Scale                    & $0.1$                          \\
    Replay Buffer Size              & $10^6$ (MT10) / $10^7$ (MT50)  \\
    Network Soft Update Rate        & $0.005$                        \\
    
    \bottomrule
  \end{tabular}
\end{table}

\begin{table}[htbp]
    \caption{Genetic Algorithm Settings}
      \label{table:evolutionary}
      \centering
      \begin{tabular}{ll}
        \toprule
        \textbf{Parameter}          & \textbf{Value}            \\ 
        \midrule
        Task Population Size        & $3$ (MT10) / $4$ (MT50)    \\
        Module Weight Precision            & $2$ (MT10) / $4$ (MT50)    \\
        Start Stages                & $3$                       \\
        Max Evaluate Times          & $2$                       \\
        Crossover Cross Population threshold & $0.9$            \\
        Crossover Rate              & $0.5$                    \\
        Mutate Cross Population threshold    & $0.95$           \\
        Mutate Best Genotype                 & $0.5$            \\
        Mutate Rate                 & $0.15$                    \\
        
        \bottomrule
      \end{tabular}
\end{table}

\end{document}